%% file: iclr2021_conference.tex
\documentclass{article} 
\usepackage{iclr2021_conference,times}

\input{math_commands.tex}

\usepackage{hyperref}
\usepackage{url}
\usepackage{booktabs}
\usepackage{enumitem}
\usepackage{graphicx}
\usepackage{subcaption}
\usepackage{wrapfig}

\title{Action Guidance: Getting the Best of Sparse Rewards and Shaped Rewards for Real-time Strategy Games}

\author{Antiquus S.~Hippocampus, Natalia Cerebro \& Amelie P. Amygdale \thanks{ Use footnote for providing further information
about author (webpage, alternative address)---\emph{not} for acknowledging
funding agencies.  Funding acknowledgements go at the end of the paper.} \\
Department of Computer Science\\
Cranberry-Lemon University\\
Pittsburgh, PA 15213, USA \\
\texttt{\{hippo,brain,jen\}@cs.cranberry-lemon.edu} \\
\And
Ji Q. Ren \& Yevgeny LeNet \\
Department of Computational Neuroscience \\
University of the Witwatersrand \\
Joburg, South Africa \\
\texttt{\{robot,net\}@wits.ac.za} \\
\AND
Coauthor \\
Affiliation \\
Address \\
\texttt{email}
}
\author{Shengyi Huang \\
College of Computing \& Informatics\\
Drexel University\\
Philadelphia, PA 19104 \\
\texttt{sh3397@drexel.edu}
\And
Santiago Onta\~{n}\'{o}n \thanks{Currently at Google} \\
College of Computing \& Informatics\\
Drexel University\\
Philadelphia, PA 19104 \\
\texttt{so367@drexel.edu}
}

\iclrfinalcopy 
\begin{document}

\maketitle

\begin{abstract}
Training agents using Reinforcement Learning in games with sparse rewards is a challenging problem, since large amounts of exploration are required to retrieve even the first reward. To tackle this problem, a common approach is to use reward shaping to help exploration. However, an important drawback of reward shaping is that agents sometimes learn to optimize the shaped reward instead of the true objective. In this paper, we present a novel technique that we call  \emph{action guidance} that successfully trains agents to eventually optimize the true objective in games with sparse rewards while maintaining most of the sample efficiency that comes with reward shaping. We evaluate our approach in a simplified real-time strategy (RTS) game simulator called $\mu$RTS. 
\end{abstract}

Training agents using Reinforcement Learning with sparse rewards is often difficult~\citep{pathakICMl17curiosity}. First, due to the sparsity of the reward, the agent often spends the majority of the training time doing inefficient exploration and sometimes not even reaching the first sparse reward during the entirety of its training.  Second, even if the agents have successfully retrieved some sparse rewards, performing proper credit assignment is challenging among complex sequences of actions that have led to theses sparse rewards. Reward shaping~\citep{ng1999policy} is a  widely-used technique designed to mitigate this problem. It works by providing intermediate rewards that lead the agent towards the sparse rewards, which are the true objective. For example, the sparse reward for a game of Chess is naturally +1 for winning, -1 for losing, and 0 for drawing, while a possible shaped reward might be +1 for every enemy piece the agent takes.  
One of the critical drawbacks for reward shaping is that the agent sometimes learns to optimize for the shaped reward instead of the real objective. Using the Chess example, the agent might learn to take as many enemy pieces as possible while still losing the game. A good shaped reward achieves a nice balance between letting the agent find the sparse reward and being too shaped (so the agent learns to just maximize the shaped reward), but this balance can be difficult to find.

In this paper, we present a novel technique called  \emph{action guidance} that successfully trains the agent to eventually optimize over sparse rewards while maintaining most of the sample efficiency that comes with reward shaping. It works by constructing a \emph{main agent} that only learns from the sparse reward function $R_{\mathcal{M}}$ and some \emph{auxiliary agents}  that learn from the shaped reward function $R_{\mathcal{A}_1}, R_{\mathcal{A}_2}, \dots, R_{\mathcal{A}_n}$. During training, we use the same rollouts to train the main and auxiliary agents and initially set a high-probability of the main agent to take \emph{action guidance} from the auxiliary agents, that is,  \emph{the main agent will execute actions sampled from the auxiliary agents}. Then the main agent and auxiliary agents are updated via off-policy policy gradient. As the training goes on, the main agent will get more independent and execute more actions sampled from its own policy. Auxiliary agents learn from shaped rewards and therefore make the training sample-efficient, while the main agent learns from the original sparse reward and therefore makes sure that the agents will eventually optimize over the true objective. 
We can see action guidance as combining reward shaping to train auxiliary agents interlieaved with a sort of imitation learning to guide the main agent from these auxiliary agents.



\begin{figure}[t]
  \centering
     \begin{subfigure}[b]{0.45\textwidth}
         \centering
         \includegraphics[width=\textwidth]{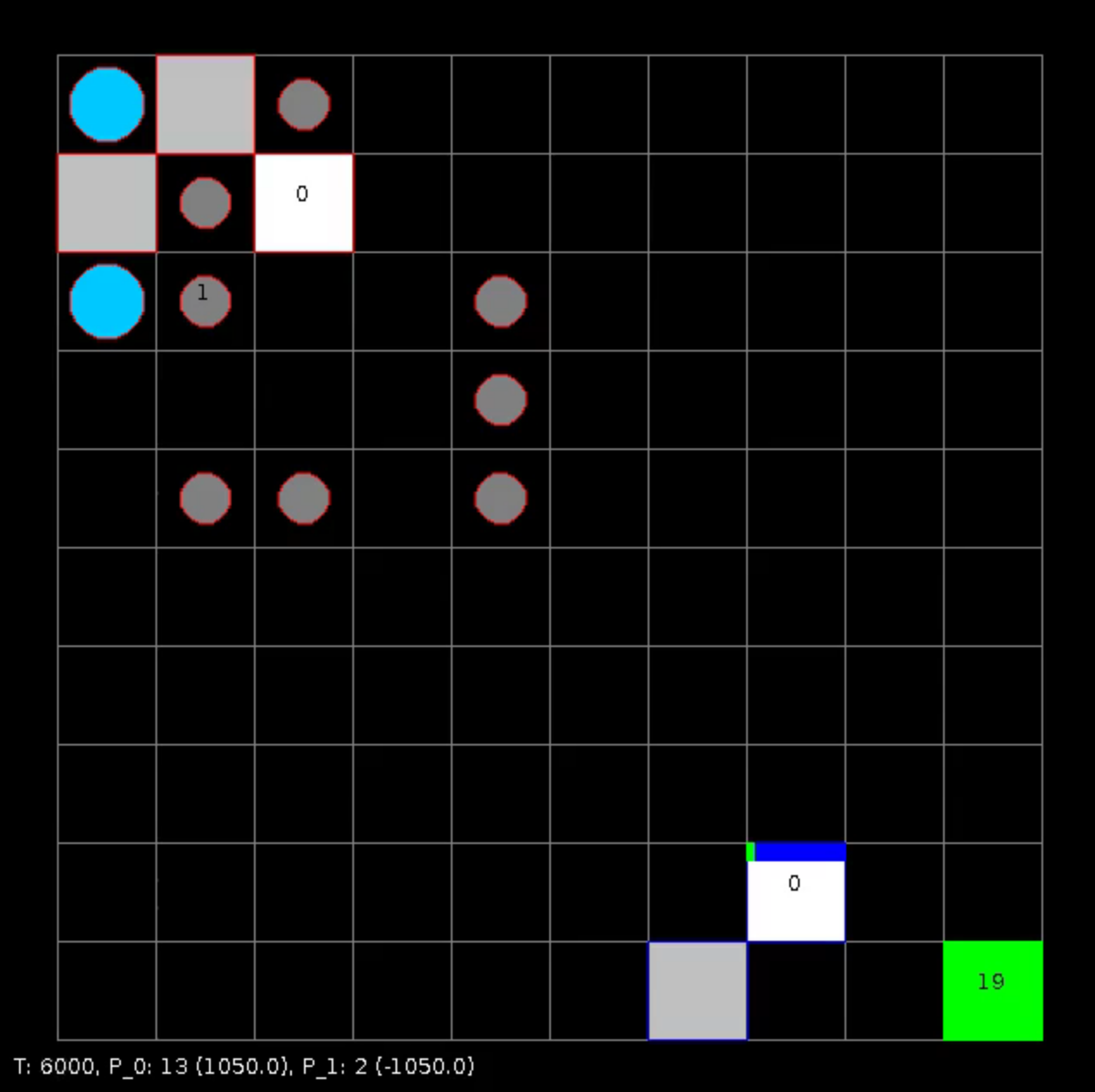}
         \caption{shaped reward\\{(\url{https://streamable.com/o797ca})}}
         \label{fig:sub:shaped_enemy}
     \end{subfigure}
     \hspace{1cm}
     \begin{subfigure}[b]{0.45\textwidth}
         \centering
         \includegraphics[width=\textwidth]{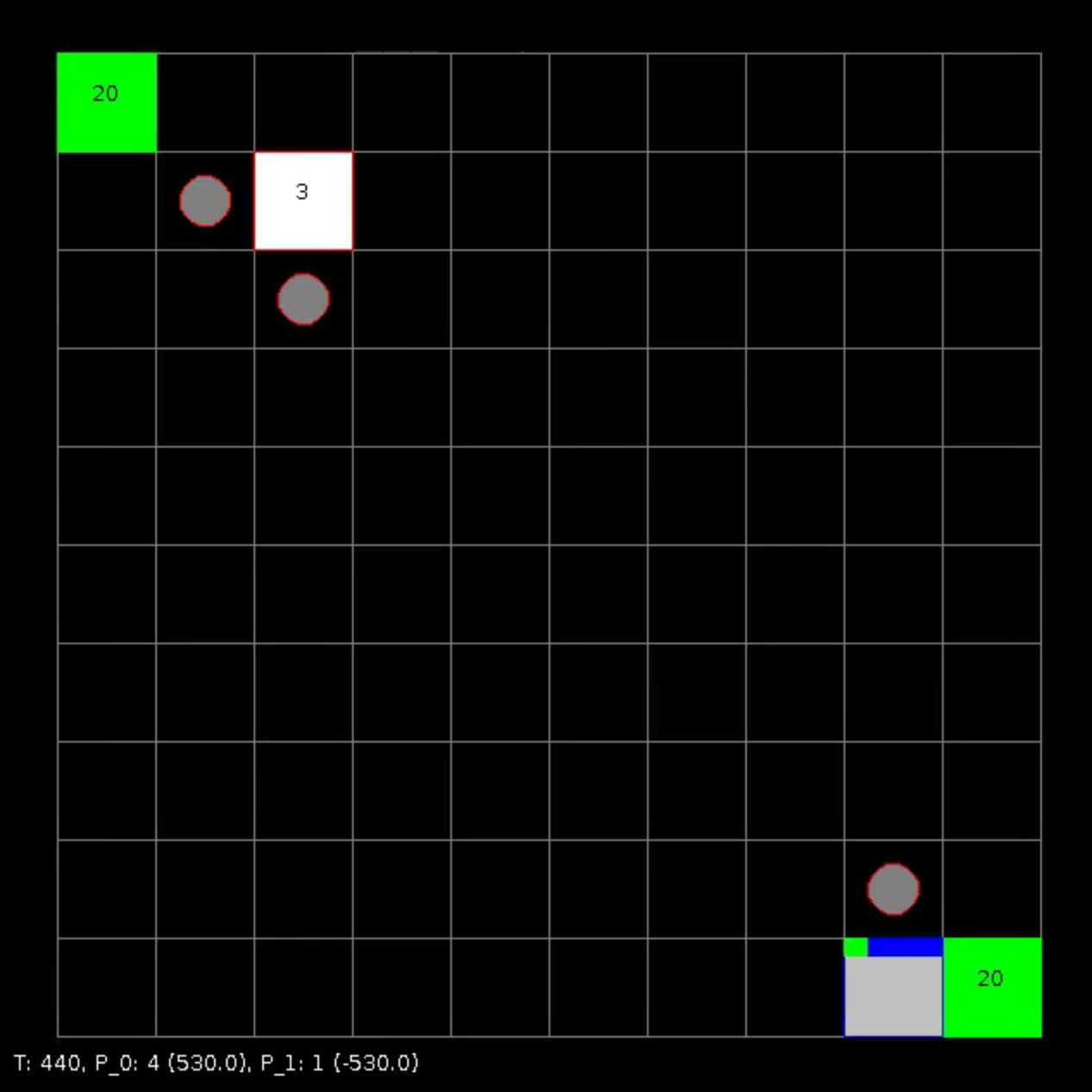}
         \caption{action guidance\\{(\url{https://streamable.com/hh7abp})}}
         \label{fig:sub:ac_enemy}
     \end{subfigure}
      \caption{The screenshot shows the typical learned behavior of agents in the task of DefeatRandomEnemy.
      (a) shows that an agent trained with some shaped reward function $R_{\mathcal{A}_1}$ learns many helpful behaviors such as building workers (grey circles), combat units (blue circles), and barracks (grey square) or using owned units (with red boarder) to attack enemy units (with blue border), but does not learn to win as fast as possible (i.e. it still does not win at internal time step $t=6000$). In contrast, (b) shows an agent trained with action guidance  optimizes over the match outcome and learns to win as fast as possible (i.e. about to win the game at $t=440$), with its main agent learning from the match outcome reward function $R_{\mathcal{M}}$   and a singular auxiliary agent learning from the same shaped reward function $R_{\mathcal{A}_1}$. Click on the link below figures to see the full videos of trained agents.}
      \label{fig:enemy}
\end{figure}

We examine action guidance in the context of a real-time strategy (RTS) game simulator called $\mu$RTS for three sparse rewards tasks of varying difficulty. For each task, we compare the performance of training agents with the sparse reward function $R_{\mathcal{M}}$, a shaped reward function $R_{\mathcal{A}_1}$, and action guidance with 
a singular auxiliary agent learning from $R_{\mathcal{A}_1}$.
The main highlights are:

\paragraph{Action guidance is sample-efficient.} Since the auxiliary agent learns from $R_{\mathcal{A}_1}$ and the main agent takes action guidance from the auxiliary agent during the initial stage of training, the main agent is more likely to discover the first sparse reward more quickly and learn more efficiently. Empirically, action guidance reaches almost the same level of sample efficiency as reward shaping in all of the three tasks tested.  
\paragraph{The true objective is being optimized.} 
During the course of training, the main agent has never seen the shaped rewards. This ensures that the main agent, which is the agent we are really interested in, is always optimizing against the true objective and is less biased by the shaped rewards.  
As an example, Figure~\ref{fig:enemy} shows that the main agent trained with action guidance eventually learns to win the game as fast as possible, even though it has only learned from the match outcome reward (+1 for winning, -1 for losing, and 0 for drawing). In contrast, the agents trained with reward shaping learn more diverse sets of behaviors which result in high shaped reward.


To support further research in this field, we make our source code available at GitHub\footnote{\url{https://github.com/anonymous-research-code/action-guidance}
}, as well as all the metrics, logs, and recorded videos\footnote{
\url{https://app.wandb.ai/vwxyzjn/action-guidance}
}. 
\section{Related Work}
In this section, we briefly summarize the popular techniques  proposed to address the challenge of sparse rewards.

{\bf Reward Shaping.} Reward shaping is a common technique where the human designer uses domain knowledge to define additional intermediate rewards for the agents. \citet{ng1999policy} show that a slightly more restricted form of state-based reward shaping has better theoretical properties for preserving the optimal policy. 

{\bf Transfer and Curriculum Learning.}  Sometimes learning the target tasks with sparse rewards is too challenging, and it is more preferable to learn some easier tasks first. \emph{Transfer learning} leverages this idea and trains agents with some easier source tasks and then later transfer the knowledge through value function~\citep{Taylor2007TransferLV} 
or reward shaping~\citep{svetlik2017automatic}.
\emph{Curriculum learning} further extends transfer learning by automatically designing and choosing a full sequences of source tasks (i.e. a curriculum)~\citep{narvekar2018learning}.

{\bf Imitation Learning.} Alternatively, it is possible to directly provide examples of human demonstration or expert replay for the agents to mimic via Behavior Cloning (BC)~\citep{bain1995framework},
which uses supervised learning to learn a policy given the state-action pairs from expert replays. Alternatively, Inverse Reinforcement Learning (IRL)~\citep{abbeel2004apprenticeship}
recovers a reward function from expert demonstrations to be used to train agents.

{\bf Curiosity-driven Learning.} Curiosity driven learning seeks to design \emph{intrinsic} reward functions~\citep{pathak18largescale} using metrics  such as prediction errors~\citep{houthooft2016curiosity}
and ``visit counts''~\citep{bellemare2016unifying,lopes2012exploration}.
These intrinsic rewards encourage the agents to explore unseen states. 

{\bf Goal-oriented Learning.} In certain tasks, it is possible to describe a goal state and use it in conjunction with the current state as input~\citep{schaul2015universal}. Hindsight experience replay (HER)~\citep{andrychowicz2017hindsight} develops better utilization of existing data in experience replay by replaying each episode with different goals. HER is shown to be an effective technique in sparse rewards tasks.

{\bf Hierarchical Reinforcement Learning (HRL).} If the target task is difficult to learn directly, it is also possible to hierarchically structure the task using experts' knowledge and train hierarchical agents, which generally involves a main agent that learns abstract goals, time, and actions, as well as auxiliary agents that learn primitive actions and specific goals~\citep{dietterich2000hierarchical}. 
HRL is especially popular in RTS games with combinatorial action spaces~\citep{pang2019reinforcement,ye2020mastering}.

The most closely related work is perhaps Scheduled Auxiliary Control (SAC-X)~\citep{riedmiller2018learning}, which is an HRL algorithm that trains auxiliary agents to perform primitive actions with shaped rewards and a main agent to schedule the use of auxiliary agents with sparse rewards. However, our approach differs in the treatment of the main agent. Instead of learning to \emph{schedule} auxiliary agents, our main agent learns to act in the entire action space by \emph{taking action guidance} from the auxiliary agents. There are two intuitive benefits to our approach since our main agent learns in the full action space. First, during policy evaluation our main agent does not have to commit to a particular auxiliary agent to perform actions for a fixed number of time steps like it is usually done in SAC-X. Second, learning in the full action space means the main agent will less likely suffer from the definition of hand-crafted sub-tasks, which could be incomplete or biased.

\section{Background}
We consider the Reinforcement Learning problem in a Markov Decision Process (MDP) denoted as $(S,A,P, \rho_0, r,\gamma, T)$, where $S$ is the state space, $A$ is the discrete action space, $P: S \times A \times S \rightarrow [0, 1]$ is the state transition probability, $\rho_0: S\rightarrow [0,1]$ is the the initial state distribution, $r: S \times A \rightarrow \mathbb{R}$ is the reward function, $\gamma$ is the discount factor, and $T$ is the maximum episode length. A stochastic policy $\pi_{\theta}: S \times A \rightarrow [0,1]$, parameterized by a parameter vector $\theta$, assigns a probability value to an action given a state. The goal is to maximize the expected discounted return of the policy:
\begin{align*}
    \mathbb{E}_{\tau}\left[\sum_{t=0}^{T-1} \gamma^{t} r_{t}\right], \begin{aligned}
        &\text { where } \tau \text { is the trajectory } \left(s_{0}, a_{0}, r_{0}, s_{1}, \dots, s_{T-1}, a_{T-1}, r_{T-1}\right)\\
        &\text { and } s_{0} \sim \rho_{0}, s_t \sim P(\cdot \vert s_{t-1}, a_{t-1}), a_t \sim \pi_{\theta}(\cdot \vert s_t), r_{t}=r\left(s_{t}, a_{t}\right)
    \end{aligned}
\end{align*}

\paragraph{Policy Gradient Algorithms.} The core idea behind policy gradient algorithms is to obtain the \textsl{policy gradient} $\nabla_{\theta}J$ of the expected discounted return with respect to the policy parameter $\theta$. Doing gradient ascent $\theta = \theta + \nabla_{\theta}J$ therefore maximizes the expected discounted reward. Earlier work proposes the following policy gradient estimate to the objective $J$~\citep{sutton2018reinforcement}:
\begin{align*}
    g_{\text{policy}, \theta} = \mathbb{E}_{\tau\sim\pi_\theta}\left[ \sum_{t=0}^{T-1} \nabla_{\theta}\log\pi_{\theta}(a_t|s_t)G_t \right]\mathrm{,} 
\end{align*}
where $G_{t} = \sum_{k=0}^{\infty} \gamma^{k} r_{t+k}$ denotes the discounted return following time $t$. This gradient estimate, however, suffers from large variance~\citep{sutton2018reinforcement} and the following gradient estimate is suggested instead:
\begin{equation*}
    g_{\text{policy}, \theta} =\mathbb{E}_{\tau}\left [ \nabla_{\theta} \sum_{t=0}^{T-1} \log\pi_{\theta}(a_t|s_t) A(\tau, V, t)\right] \mathrm{,} 
\end{equation*}
where $A(\tau, V, t)$ is the General Advantage Estimation (GAE)~\citep{schulman2015high}, which measures ``how good is $a_t$ compared to the usual actions'', and $V: S\rightarrow\mathbb{R}$ is the state-value function.

\section{Action Guidance}


The key idea behind {\em action guidance} is to create a main agent that trains on the sparse rewards, and creating some auxiliary agents that are trained on shaped rewards. During the initial stages of training, the main agent has a high probability to take \emph{action guidance} from the auxiliary agents, that is, the main agent can execute actions sampled from the auxiliary agents, rather than from its own policy. As the training goes on, this probability decreases, and the main agent  executes more actions sampled from its own policy. During training, the main and auxiliary agents are updated via off-policy policy gradient. Our use of auxiliary agents makes the training sample-efficient, and our use of the main agent, who only sees its own sparse reward, makes sure that the agent will eventually optimize over the true objective of sparse rewards. In a way, {\em action guidance} can be seen as training agents using shaped rewards, while having the main agent learn by imitating from them.

Specifically, let us define $\mathcal{M}$ as the MDP that the main agent learns from and $\mathcal{A}=\{\mathcal{A}_1, \mathcal{A}_2, ..., \mathcal{A}_k\}$ be a set of auxiliary MDPs that the auxiliary agents learn from. In our constructions, $\mathcal{M}$ and $\mathcal{A}$ share the same state, observation, and action space. However, the reward function for $\mathcal{M}$ is $R_{\mathcal{M}}$, which is the sparse reward function, and reward functions for $\mathcal{A}$ are $ R_{\mathcal{A}_1}, ..., R_{\mathcal{A}_k}$, which are the shaped reward functions. For each of these MDPs $\mathcal{E} \in \mathcal{S}=\{\mathcal{M}\} \cup\mathcal{A}$ above, let us initialize a policy $\pi_{\theta_\mathcal{E}}$ parameterized by parameters $\theta_\mathcal{E}$, respectively. Furthermore, let us use $\pi_\mathcal{S} = \{\pi_{\theta_\mathcal{E}} |\mathcal{E} \in \mathcal{S}\}$ to denote the set of these initialized policies.

At each timestep $t$, let us use some exploration strategy $S$ that selects a policy $\pi_b \in \pi_\mathcal{S}$ to sample an action $a_t$ given $s_t$.  At the end of the episode, each policy $\pi_{\theta}    \in  \pi_\mathcal{S}$  can be updated via its off-policy policy gradient~\citep{degris2012off,levine2020offline}:
\begin{align}
    \mathbb{E}_{\tau\sim\pi_{\theta_b}}\left[\left(\prod_{t=0}^{T-1} \frac{\pi_{\theta} \left(a_t | s_t\right)}{\pi_{\theta_b}\left(a_t | s_t\right)}\right) \sum_{t=0}^{T-1}  \nabla_{\theta} \log \pi_{\theta} \left(a_t | s_t\right) A(\tau, V, t) \right]\label{eq:off-gradient}
\end{align}
When $\pi_{\theta} =\pi_{\theta_b}$, the gradient in Equation~\ref{eq:off-gradient} means on-policy policy gradient update for $\pi_{\theta} $. Otherwise, the objective means off-policy policy gradient update for $\pi_{\theta} $. 

\subsection{Practical Algorithm}
The gradient in Equation~\ref{eq:off-gradient} is unbiased, but its product of importance sampling ratio $\left(\prod_{t=0}^{T-1} \frac{\pi_{\theta} \left(a_t | s_t\right)}{\pi_{\theta_b}\left(a_t | s_t\right)}\right)$ is known to cause high variance~\citep{wang2016sample}. In practice, we clip the gradient the same way as Proximal Policy Gradient (PPO)~\citep{schulman2017proximal}:
\begin{align}
  L^{CLIP}(\theta)\ = \mathbb{E}_{\tau\sim\pi_{\theta_b}}\left[\sum_{t=0}^{T-1}\left[\nabla_{\theta}
        \mathrm{min} \left(
            \rho_t(\theta)  A(\tau, V, t),
            \operatorname{clip}\left(\rho_t(\theta), \varepsilon \right) A(\tau, V, t) \right) \right]
        \right] \label{eq:real-objective}\\ \rho_t(\theta)  = \frac{\pi_{\theta}\left(a_t | s_t\right)}{\pi_{\theta_b}\left(a_t | s_t\right)}, \quad \operatorname{clip}\left(\rho_t(\theta), \varepsilon\right) = \begin{cases}
1-\varepsilon &\text{if } \rho_t(\theta) < 1-\varepsilon\\
1+\varepsilon &\text{if } \rho_t(\theta) > 1+\varepsilon\\
\rho_t(\theta) &\text{otherwise}\nonumber
\end{cases}
\end{align}
During the optimization phase, the agent also learns the value function and  maximize the policy's entropy. We therefore optimize the following joint objective for each $\pi_{\theta}    \in  \pi_\mathcal{S}$:
\begin{align}
    L^{CLIP}(\theta) = L^{CLIP}(\theta) - c_1L^{VF}(\theta) + c_2S[\pi_{\theta_b}], \label{eq:full-objective} 
\end{align}
where  $c_1, c_2$ are coefficients, $S$ is an entropy bonus, and $L^{VF}$ is the squared error loss for the value function associated with $\pi_{\theta}$ as done by \cite{schulman2017proximal}.
Although action guidance can be configured to leverage multiple auxiliary agents that learn diversified reward functions, we only use one auxiliary agent for the simplicity of experiments.
In addition, we use $\epsilon$-greedy as the exploration strategy $S$ for determining the behavior policy. That is, at each timestep $t$, the behavior policy is selected to be $\pi_{\theta_\mathcal{M}}$ with probability $1-\epsilon$  and $\pi_{\theta_\mathcal{D}}$ for $\mathcal{D}\in \mathcal{A}$ with probability $\epsilon$ (note that is $\epsilon$ is different from the clipping coefficient $\varepsilon$ of PPO). Additionally, $\epsilon$ is set to be a constant $0.95$ at start for some period of time steps (e.g. 800,000), which we refer to as the \emph{shift} period (the time it takes to start ``shifting'' focus away from the auxiliary agents), then it is set to linearly decay to $\epsilon_{end}$ for some period of time steps (e.g. 1,000,000), which we refer to as the \emph{adaptation} period (the time it takes for the main agent to fully ``adapt'' and become more independent).

\subsection{Positive Learning Optimization}
During our initial experiments, we found the main agent sometimes did not learn useful policies. Our hypothesis is that this was because the main agent is updated with too many trajectories with zero reward. Doing a large quantities of updates of these zero-reward trajectories actually causes the policy to converge prematurely, which is manifested by having low entropy in the action probability distribution. 

To mitigate this issue of having too many zero-reward trajectories, we use a preliminary  code-level optimization called Positive Learning Optimization (PLO). After collecting the rollouts, PLO works by skipping the gradient update for $\pi_{\theta_\mathcal{E}} \in  \pi_\mathcal{S}$ and its value function if the rollouts contains no reward according to $R_{\mathcal{E}}$. Intuitively, PLO makes sure that the main agent learns from meaningful experience that is associated with positive rewards. To confirm its effectiveness, we provide an ablation study of PLO in the experiment section.


\section{Evaluation}
\label{sec:evaluation_environment}

We use $\mu$RTS\footnote{\url{https://github.com/santiontanon/microrts}} as our testbed, which is a minimalistic RTS game maintaining the core features that make RTS games challenging from an AI point of view: simultaneous and durative actions, large branching factors and real-time decision making. To interface with $\mu$RTS, we use gym-microrts\footnote{\url{https://github.com/vwxyzjn/gym-microrts}}~\citep{huang2020closer} to conduct our experiments. The details of gym-microrts as a RL interface can be found at Appendix~\ref{sec:details_on_murts}. 

\subsection{Tasks Description}
We examine the three following sparse reward tasks with a range of difficulties. For each task, we compare the performance of training agents with the sparse reward function $R_{\mathcal{M}}$, a shaped reward function $R_{\mathcal{A}_1}$, and action guidance with a single auxiliary agent learning from $R_{\mathcal{A}_1}$. Here are the descriptions of these environments and their reward functions.
\begin{enumerate}
    \item LearnToAttack: In this task, the agent's objective is to learn move to the other side of the map where the enemy units live and start attacking them.  
    Its $R_{\mathcal{M}}$ gives a +1 reward for each valid attack action the agent issues. This is of sparse reward because the action space is so large: the agent could have build a barracks or produce a unit; it is unlikely that the agents will by chance issue lots of moving actions (out of 6 action types) with correct directions (out of 4 directions) and then start attacking. Its $R_{\mathcal{A}_1}$ gives the difference between previous and current Euclidean distance between the enemy base and its closet unit owned by the agent as the shaped reward in addition to  $R_{\mathcal{M}}$.
    \item ProduceCombatUnits: In this task, the agent's objective is to learn to build as many combat units as possible. Its $R_{\mathcal{M}}$ gives a +1 reward for each combat unit the agent produces. This is a more challenging task because the agent needs to learn 1) harvest resources, 2) produce barracks, 3) produce combat units once enough resources are gathered, 4) move produced combat units out of the way so as to not block the production of new combat units. Its $R_{\mathcal{A}_1}$ gives +1 for constructing every building (e.g. barracks), +1 for harvesting resources, +1 for returning resources, and +7 for each combat unit it produces.
    \item DefeatRandomEnemy: In this task, the agent's objective is to defeat a biased random bot of which the attack, harvest and return actions have 5 times the probability of other actions. Additionally, the bot subjects to the same gym-microrts' limitation (See Appendix~\ref{sec:limitations}) as the agents used in our experiments. Its $R_{\mathcal{M}}$ gives the match outcome as the reward (-1 on a loss, 0 on a draw and +1 on a win). This is the most difficult task we examined because the agent is subject to the full complexity of the game, being required to make both macro-decisions (e.g. deciding the high-level strategies to win the game) and micro-decisions (e.g. deciding which enemy units to attack. In comparison, its $R_{\mathcal{A}_1}$ gives +5 for winning, +1 for harvesting one resource, +1 for returning resources, +1 for producing one worker, +0.2 for constructing every building, +1 for each valid attack action it issues, +7 for each combat unit it produces, and $+(0.2*d)$ where $d$ is difference between previous and current Euclidean distance between the enemy base and its closet unit owned by the agent.

\end{enumerate}

\subsection{Agent Setup}
We use PPO~\citep{schulman2017proximal} as the base DRL algorithm to incorporate action guidance. The details of the implementation, neural network architecture, hyperparameters, proper handling of $\mu$RTS's action space and invalid action masking~\citep{huang2020closer} can be found in Appendix~\ref{sec:details_on_ppo}.
We compared the following strategies:
\begin{enumerate}
    \item \textbf{Sparse reward (first baseline).} This agent is trained with PPO on $R_{\mathcal{M}}$ for each task.
    \item \textbf{Shaped reward (second baseline).} This agent is trained with PPO on $R_{\mathcal{A}_1}$ for each task.
    \item \textbf{Action guidance~-~long adaptation.} The agent is trained with PPO + action guidance with $\mathit{shift}=2,000,000$  time steps, $\mathit{adaptation}=7,000,000$  time steps, and $\epsilon_{end}= 0.0$
    \item \textbf{Action guidance~-~short adaptation.} The agent is trained with PPO + action guidance with $\mathit{shift}=800,000$  time steps, $\mathit{adaptation}=1,000,000$  time steps, and $\epsilon_{end}= 0.0$
    \item \textbf{Action guidance~-~mixed policy.} The agent is trained with PPO + action guidance with $\mathit{shift}=2,000,000$  time steps and $\mathit{adaptation}=2,000,000$  time steps, and $\epsilon_{end}= 0.5$. We call this agent ``mixed policy'' because it will eventually have 50\% chance to sample actions from the main agent and 50\% chance to sample actions from the auxiliary agent. It is effectively having mixed agent making decisions jointly.
\end{enumerate}
Although it is desirable to add SAC-X to the list of strategies compared, it was not designed to handle domains with large discrete action spaces. 
Lastly, we also toggle the PLO option for action guidance~-~long adaptation, action guidance~-~short adaptation, action guidance~-~mixed policy, and sparse reward training strategies for a preliminary ablation study.

\subsection{Experimental Results}
Each of the 6 strategies is evaluated in 3 tasks with 10 random seeds. We report the results in Table~\ref{tab:results}. All the learning curves can be found in Appendix~\ref{sec:learningcurves}. Below are our observations.

\begin{figure}[t]
\includegraphics[width=0.49\columnwidth]{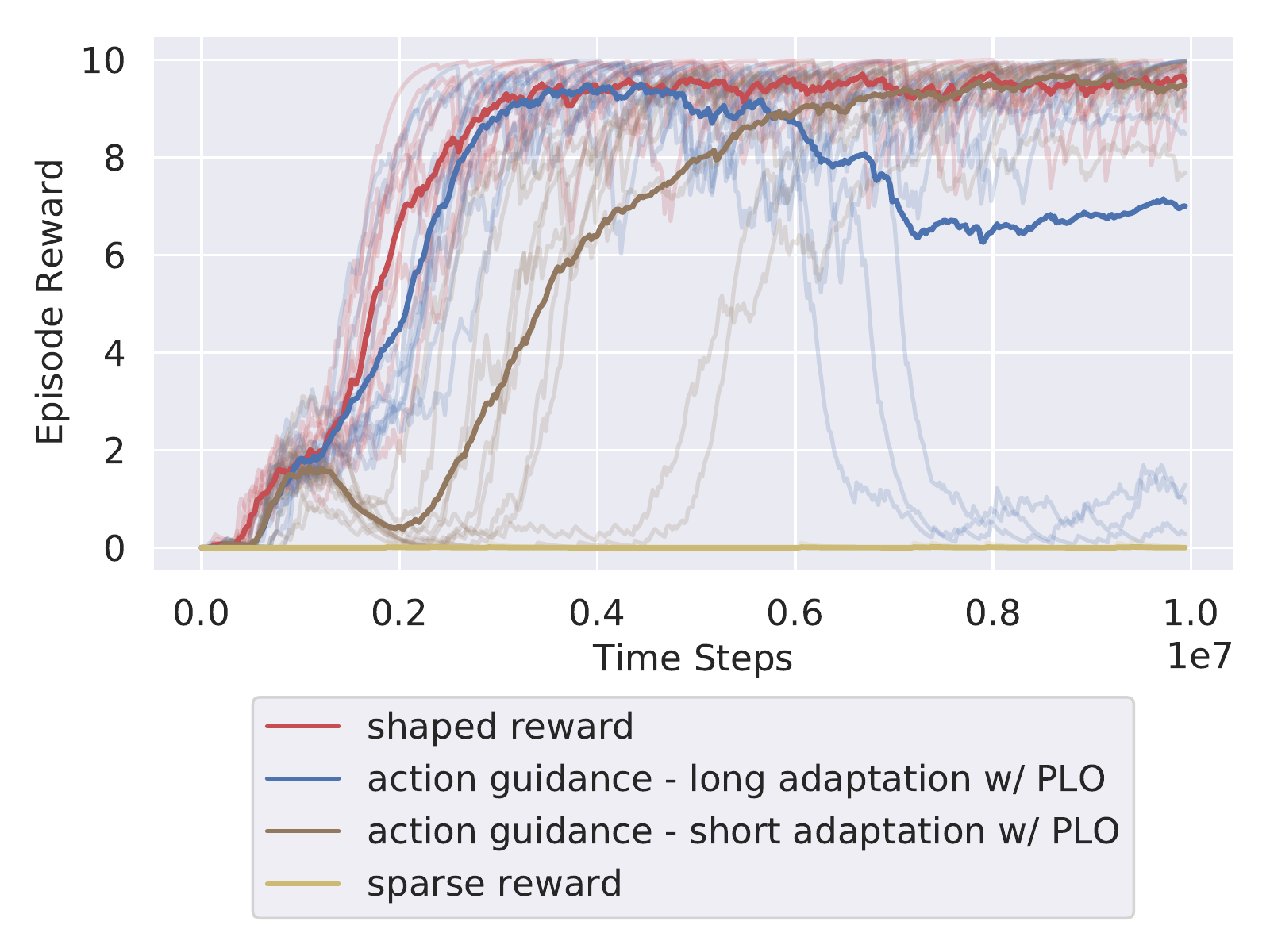}
\includegraphics[width=0.49\columnwidth]{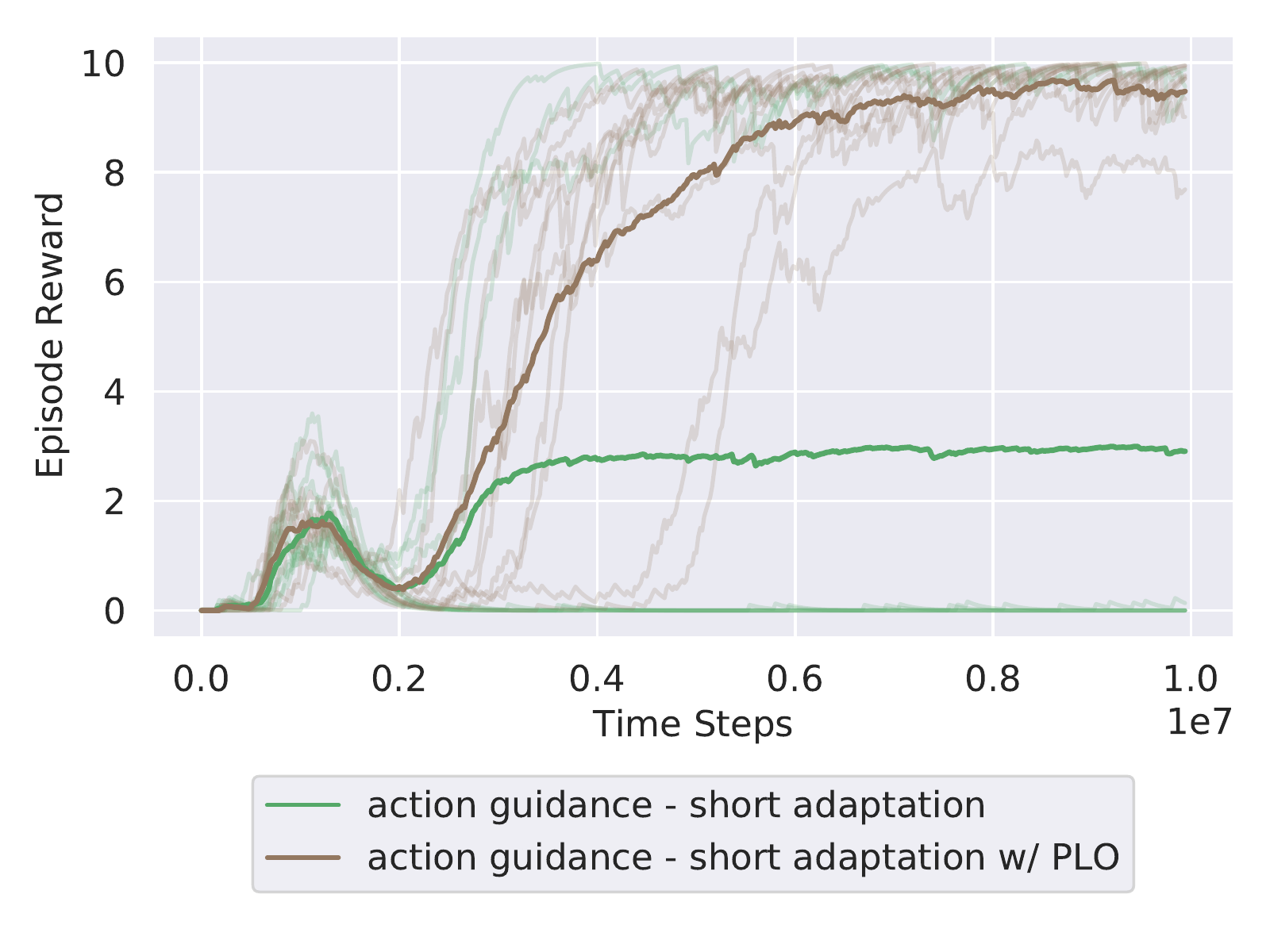}
  \caption{The faint lines are the actual episode reward of each seed for selected strategies in ProduceCombatUnits; solid lines are their means. The left figure showcase the sample-efficiency of action guidance; the right figure is a motivating example for PLO.}
  \vspace{-0.2cm}
  \label{fig:ProduceCombatUnits}
\end{figure}

\textbf{Action guidance is almost as sample-efficient as reward shaping.}
Since the auxiliary agent learns from $R_{\mathcal{A}_1}$ and the main agent takes a lot of action guidance from the auxiliary agent during the shift period, the main agent is more likely to discover the first sparse reward more quickly and learn more efficiently. As an example, Figure~\ref{fig:ProduceCombatUnits} demonstrates such sample-efficiency in ProduceCombatUnits, where the agents trained with sparse reward struggle to obtain the very first reward. In comparison, most action guidance related agents are able to learn almost as fast as the agents trained with shaped reward.

\begin{table*}[]
    \centering
    \caption{The average episode reward (according to $R_{\mathcal{M}}$) achieved by each training strategy in each task over 10 random seeds.}
    \label{tab:results}
\begin{small}

\begin{tabular}{llll}
\toprule
 & LearnToAttack & ProduceCombatUnit & DefeatRandomEnemy \\
\midrule
sparse reward   (first baseline)          &          3.30 $\pm$ 5.04 &                     0.00 $\pm$ 0.01 &               -0.07 $\pm$ 0.03 \\
sparse reward w/ PLO                      &          0.00 $\pm$ 0.00 &                     0.00 $\pm$ 0.01 &               -0.05 $\pm$ 0.03 \\
shaped reward  (second baseline)          &         10.00 $\pm$ 0.00 &                     \textbf{9.57 $\pm$ 0.30} &                0.08 $\pm$ 0.17 \\
action guidance - long adaptation         &         \textbf{11.00 $\pm$ 0.00} &                     8.31 $\pm$ 2.62 &                0.11 $\pm$ 0.35 \\
action guidance - long adaptation w/ PLO  &         \textbf{11.00 $\pm$ 0.01} &                     6.96 $\pm$ 4.04 &                \textbf{0.52 $\pm$ 0.35} \\
action guidance - mixed policy            &         \textbf{11.00 $\pm$ 0.00} &                     \textbf{9.67 $\pm$ 0.17} &               \textbf{0.40 $\pm$ 0.37} \\
action guidance - mixed policy  w/ PLO    &         \textbf{10.67 $\pm$ 0.12} &                     \textbf{9.36 $\pm$ 0.35} &                \textbf{0.30 $\pm$ 0.42} \\
action guidance - short adaptation        &         \textbf{11.00 $\pm$ 0.01} &                     2.95 $\pm$ 4.48 &               -0.06 $\pm$ 0.04 \\
action guidance - short adaptation w/ PLO &         \textbf{11.00 $\pm$ 0.00} &                    \textbf{9.48 $\pm$ 0.51} &               -0.05 $\pm$ 0.03 \\
\bottomrule
\end{tabular}
\vspace{-0.3cm}
\end{small}
\end{table*}
\textbf{Action guidance eventually optimizes the sparse reward.} This is perhaps the most important contribution of our paper. Action guidance eventually optimizes the main agent over the true objective, rather than optimizing shaped rewards. Using the ProduceCombatUnits task as an example, the agent trained with shaped reward would only start producing combat units once all the resources have been harvested, probably because the +1 reward for harvesting and returning resources are easy to retrieve and therefore the agents exploit them first.   Only after these resources are exhausted would the agents start searching for other sources of rewards then learn producing combat units.  

In contrast, the main agent of action guidance~-~short adaptation w/ PLO are initially guided by the shaped reward agent during the shift period. During the adaptation period, we find the main agent starts to optimize against the real objective by producing the first combat unit as soon as possible. This disrupts the behavior learned from the auxiliary agent and thus cause a visible degrade in the main agent's performance during 1M and 2M timesteps as shown in Figure~\ref{fig:ProduceCombatUnits}. As the adaption period comes to an end, the main agent becomes fully independent and learn to produce combat units and harvest resources concurrently. This behavior matches the common pattern observed in professional RTS game players and is obviously more desirable because should the enemy attack early, the agent will have enough combat units to defend.

In the DefeatRandomEnemy task, the agents trained with shaped rewards learn a variety of behaviors; some of them learn to do a worker rush while others learn to focus heavily on harvesting resources and producing units. This is likely because the agents could get similar level of shaped rewards despite having diverse set of behaviors. In comparison, the main agent of action guidance - long adaptation w/ PLO would start optimizing the sparse reward after the shift period ends; it almost always learns to do a worker rush, which an efficient way to win against a random enemy as shown in Figure~\ref{fig:enemy}.

\textbf{The hyper-parameters adaptation and shift matter.} Although the agents trained with action guidance~-~short adaptation w/ PLO learns the more desirable behavior, they perform considerably worse in the harder task of DefeatRandomEnemy. It suggests the harder that task is perhaps the longer adaptation should be set. However, in ProduceCombatUnits, agents trained with action guidance~-~long adaptation  w/ PLO exhibits the same category of behavior as agents trained with shaped reward, where the agent would only start producing combat units once all the resources have been harvested. A reasonable explanation is that higher adaptation gives more guidance to the main agents to consistently find the sparse reward, but it also inflicts more bias on how the task should be accomplished; lower adaption gives less guidance but increase the likelihood for the main agents to find better ways to optimize the sparse rewards.

\begin{figure}[t]
  \centering
     \begin{subfigure}[b]{0.45\textwidth}
         \centering
         \includegraphics[width=\textwidth]{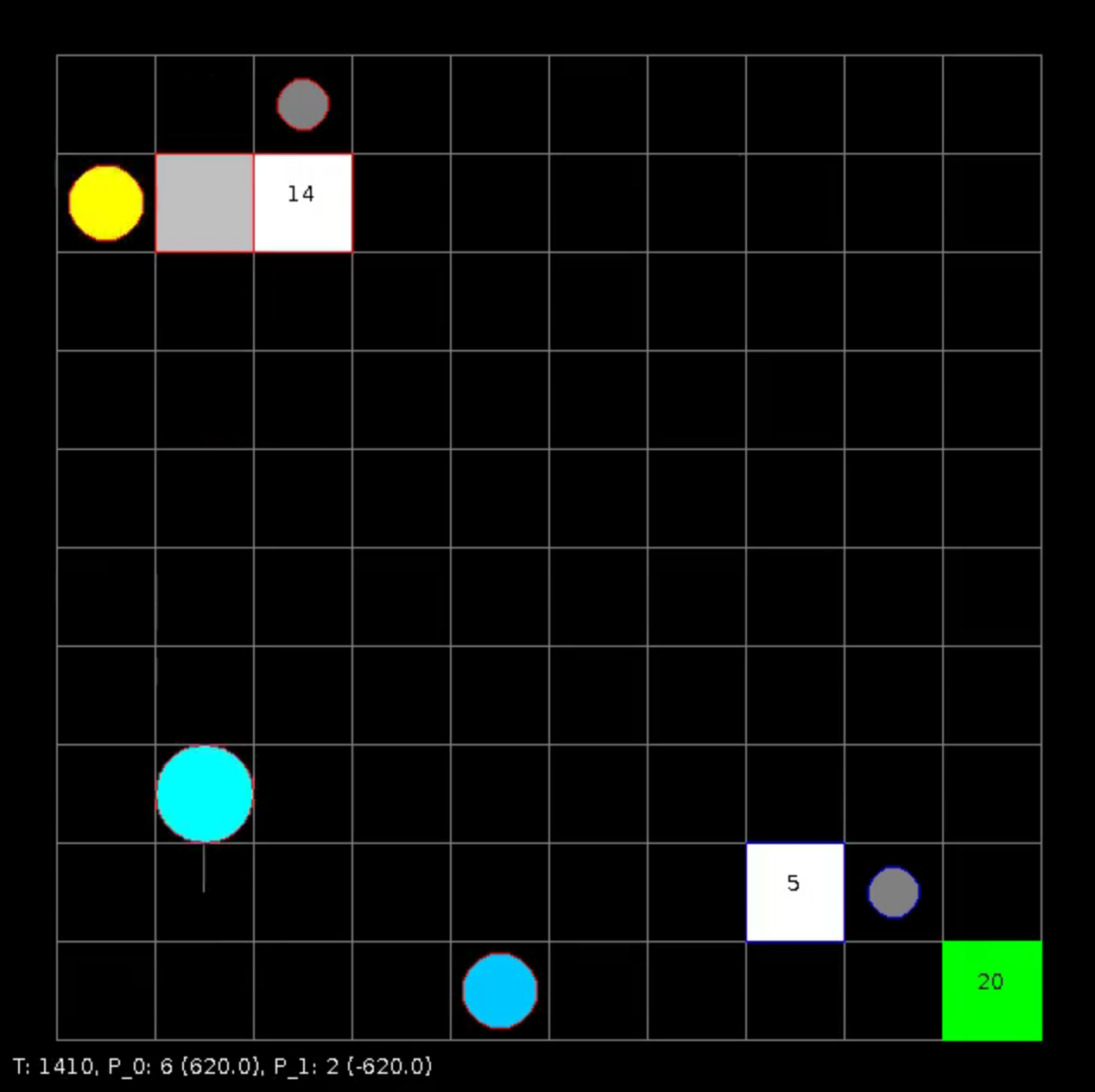}
         \caption{shaped reward\\{(\url{https://streamable.com/ytpt7u})}}
         \label{fig:sub:shaped_combatunits}
     \end{subfigure}
     \hspace{1cm}
     \begin{subfigure}[b]{0.45\textwidth}
         \centering
         \includegraphics[width=\textwidth]{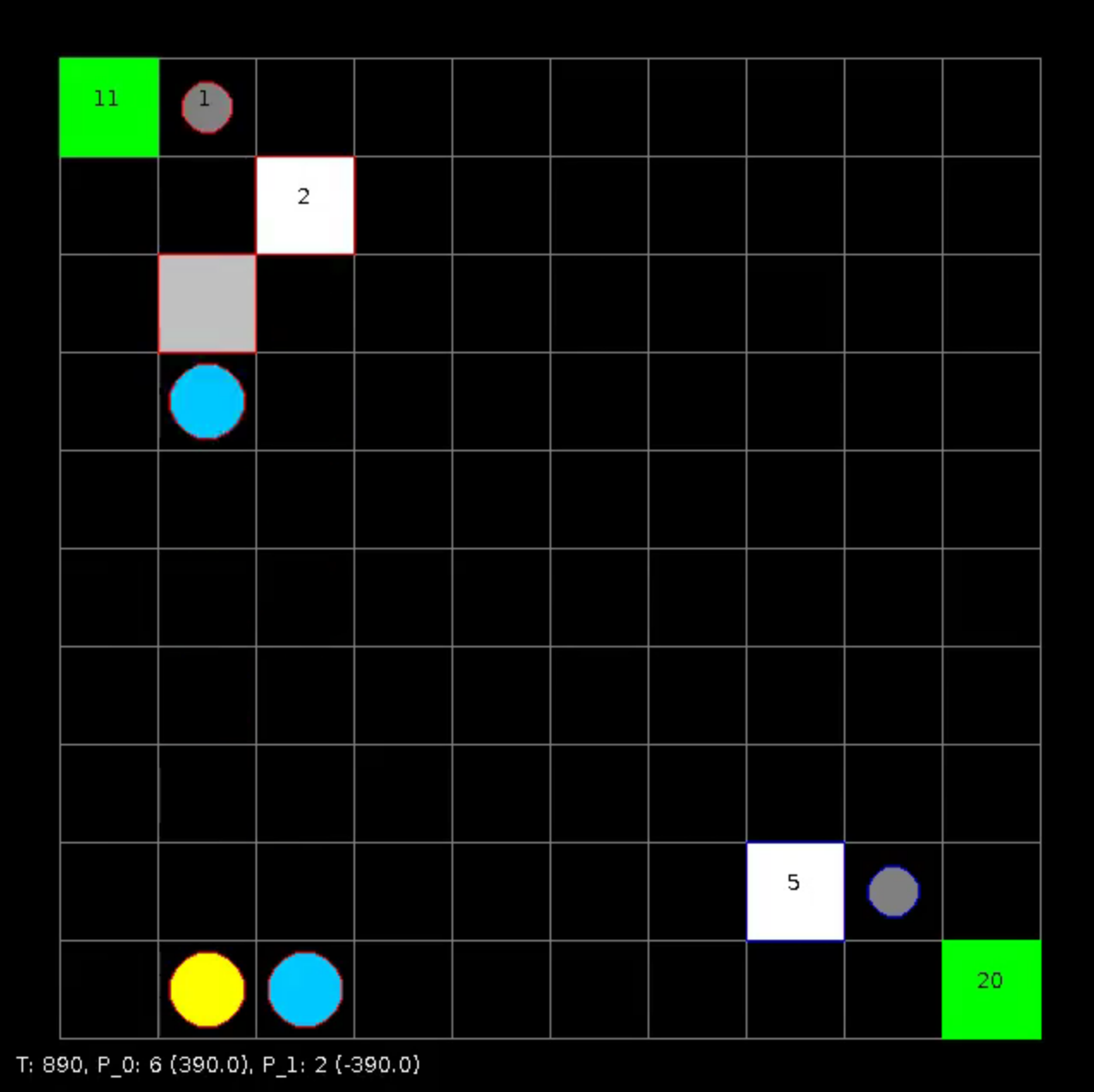}
         \caption{action guidance \\{(\url{https://streamable.com/mpzxef})}}
         \label{fig:sub:action_guidance_combatunits}
     \end{subfigure}
       \caption{The screenshot shows the typical learned behavior of agents in the task of ProduceCombatUnits. (a) shows an agent trained with shaped reward function $R_{\mathcal{A}_1}$ learn to only produce combat units once the resources are exhausted (i.e. it produces three combat units at $t=1410$). In contrary, (b) shows an agent trained with action guidance  learn to produce units and harvest resources concurrently (i.e. it produces three combat units at $t=890$). Click on the link below figures to see the full videos of trained agents.}
       \vspace{-0.2cm}
\end{figure}

\textbf{Positive Learning Optimization results are inconclusive.} We found PLO to be an interesting yet sometimes effective optimization in stabilizing the performance for agents trained with action guidance. As a motivating example, Figure~\ref{fig:ProduceCombatUnits} showcases the actual episode reward of 10 seeds in ProduceCombatUnits, where agents trained with action guidance - short adaptation and PLO seem to always converge while agents trained without PLO would only sometimes converge. However, PLO does not always help. For example, PLO actually hurt the performance of action guidance~-~long adaptation in ProduceCombatUnits by having a few degenerate runs as shown in Figure~\ref{fig:ProduceCombatUnits}. It is also worth noting the PLO does not help the sparse reward agent at all, suggesting PLO is a an optimization somewhat unique to action guidance.

\textbf{Action guidance - mixed policy is viable.} 
According to Table~\ref{tab:results}, agents trained with action guidance~-~mixed policy with or without PLO seem to perform relatively well in all three tasks examined. 
This is a interesting discovery because it suggests action guidance could go both ways: the auxiliary agents could also benefit from the learned policies of the main agent. An alternative perspective is to consider the main agent and the auxiliary agents as a whole entity that mixes different reward functions, somehow making joint decision and collaborating to accomplish common goals.




\section{Conclusions}

In this paper, we present a novel technique called  \emph{action guidance} that successfully trains the agent to eventually optimize over sparse rewards yet does not lose the sample efficiency that comes with reward shaping, effectively getting the best of both worlds. Our experiments with DefeatRandomEnemy in particular show it is possible to train a main agent on the full game of $\mu$RTS using only the match outcome reward, which suggests action guidance could serve as a promising alternative to the training paradigm of AlphaStar~\citep{vinyals2019grandmaster} that uses supervised learning with human replay data to bootstrap an  agent.
%
As part of our future work, we would like to scale up the approach to defeat stronger opponents.



\bibliography{iclr2021_conference}
\bibliographystyle{iclr2021_conference}

\clearpage

\appendix
\section*{Appendices}
\addcontentsline{toc}{section}{Appendices}
\renewcommand{\thesubsection}{\Alph{subsection}}

\subsection{Details on $\mu$RTS as a Reinforcement Learning Environment}
\label{appendix:additional-detals-murts}
For all of our experiments in this paper, we use the \emph{gym-microrts}~\citep{huang2019comparing} library which provides a reinforcement learning interface of $\mu$RTS similar to OpenAI Gym's interface~\citep{brockman2016openai}. In this section, we hope to provide details on the implementation of \emph{gym-microrts} as well as its limitations.

\subsubsection{Observation and Action Space of $\mu$RTS}
Here is a description of \emph{gym-microrts}'s observation and action space:
\label{sec:details_on_murts}
\begin{itemize}[leftmargin=*]
    \item \textbf{Observation Space.} Given a map of size $h\times w$, the observation is a tensor of shape $(h, w, n_f)$, where $n_f$ is a number of feature planes that have binary values. The observation space used in this paper uses 27 feature planes as shown in Table~\ref{tab:action-components}. A feature plane can be thought of as a concatenation of multiple one-hot encoded features. As an example, if there is a worker with hit points equal to 1, not carrying any resources, owner being Player 1, and currently not executing any actions, then the one-hot encoding features will look like the following:
    \begin{small}
        \begin{align*}
            [0,1,0,0,0],  [1,0,0,0,0],  [1,0,0], \\ [0,0,0,0,1,0,0,0],  [1,0,0,0,0,0]
        \end{align*}
    \end{small}
    The 27 values of each feature plane for the position in the map of such worker will thus be:
    \begin{small}
    \[[0,1,0,0,0,1,0,0,0,0,1,0,0,0,0,0,0,1,0,0,0,1,0,0,0,0,0]\]
    \end{small}
    \item \textbf{Action Space.} Given a map of size $h\times w$, the action is an 8-dimensional vector of discrete values as specified in Table~\ref{tab:action-components}. The first component of the action vector represents the unit in the map to issue actions to, the second is the action type, and the rest of components represent the different parameters different action types can take. 
    
\end{itemize}

\subsubsection{Limitations}
\label{sec:limitations}
In general, there are two limitations associated with \emph{gym-microrts}'s reinforcement learning interface that could hinder the trained agents from competing against existing $\mu$RTS bots. As a result, the agent trained  in this paper would subject to these limitations.

\paragraph{Frame-skipping}
Each action in $\mu$RTS takes some internal game time, measured in ticks, for the action to be completed. \emph{gym-microrts} sets the time of performing harvest action, return action, and move action to be 10 game ticks. Once an action is issued to a particular unit, the unit would be considered as a ``busy'' unit and would take additional 9 game ticks for the actions to be finalized. To speed up training, \emph{gym-microrts} by default performs frame skipping of 9 frames such that from the agent's perspective, once it executes the harvest action, return action, or move action given the current observation, those actions would be finished in the next observation. 

\paragraph{Limited Unit Action per Tick}
In general, the game engine of $\mu$RTS allows the bots to issue actions to as many units as the bots own at each game tick. This means the action space will have ``varied size'' depending on the number of units available for control. For simplicity, \emph{gym-microrts} only allows the agent to issue one action to one unit at each tick.

\subsection{Details on the Training Algorithm Proximal Policy Optimization}
\label{sec:details_on_ppo}
The DRL algorithm that we use to train the agent is Proximal Policy Optimization (PPO)~\cite{schulman2017proximal}, one of the state of the art algorithms available. The hyper-parameters of our experiments can be found in Table~\ref{tab:params}. There are three important  details regarding our PPO implementation that warrants explanation. The first detail concerns how to generate an action in the \verb MultiDiscrete  action space as defined in the OpenAI Gym environment~\citep{brockman2016openai} of gym-microrts~\citep{huang2019comparing}, the second details involves the implementation of invalid action masking~\citep{vinyals2017starcraft,Berner2019Dota2W,huang2020closer} with PPO, and the third detail is about the various code-level optimizations utilized to augment performance. As pointed out by Engstrom, Ilyas, et al.~\citep{engstrom2019implementation}, such code-level optimizations could be critical to the performance of PPO.

\begin{table*}[t]
\centering
\caption{The descriptions of observation features and action components.}
\begin{small}
\begin{tabular}{ lll} 
\toprule
Observation Features  & Planes & Description \\
\midrule
Hit Points & 5 & 0, 1, 2, 3, $\geq 4$  \\ 
Resources & 5 & 0, 1, 2, 3, $\geq 4$  \\ 
Owner &3 & player 1, -, player 2 
\\ 
Unit Types &8 & -, resource, base, barrack,worker, light, heavy, ranged \\ 
Current Action &6& -, move, harvest, return, produce, attack\\ 
\midrule
Action Components  & Range & Description \\
\midrule
Source Unit & $[0,h \times w-1]$ & the location of unit selected to perform an action  \\ 
Action Type & $[0,5]$ & NOOP, move, harvest, return, produce, attack  \\ 
Move Parameter & $[0,3]$ & north, east, south, west \\ 
Harvest Parameter & $[0,3]$  & north, east, south, west  \\
Return Parameter & $[0,3]$ & north, east, south, west  \\
Produce Direction Parameter & $[0,3]$ & north, east, south, west  \\
Produce Type Parameter & $[0,5]$ & resource, base, barrack, worker, light, heavy, ranged \\
Attack Target Unit & $[0,h\times w-1]$  & the location of unit that  will be attacked \\
\bottomrule
\end{tabular}
\end{small}
\label{tab:action-components}
\end{table*}

\subsubsection{Multi Discrete Action Generation}
To perform an action $a_t$ in $\mu$RTS, according to Table~\ref{tab:action-components}, we have to select a Source Unit, Action Type, and its corresponding action parameters. So in total, there are $hw\times6\times4\times4\times4\times4\times6\times hw = 9216(hw)^2 $  number of possible discrete actions (including invalid ones), which grows exponentially as we increase the map size. If we apply the PPO directly to this discrete action space, it would be computationally expensive to generate the distribution for $9216(hw)^2 $ possible actions. To simplify this combinatorial action space, \verb openai/baselines ~\cite{baselines} library proposes an idea to consider this discrete action to be composed from some smaller \emph{independent} discrete actions. Namely, $a_t$ is composed of smaller actions 
\begin{align*}
    &a_{t}^{\text{Source Unit}},a_{t}^{\text{Action Type}},a_{t}^{\text{Move Parameter}},a_{t}^{\text{Harvest Parameter}}, \\
    &a_{t}^{\text{Return Parameter}},a_{t}^{\text{Produce Direction Parameter}}, a_{t}^{\text{Produce Type Parameter}},a_{t}^{\text{Attack Target Unit}}
\end{align*}
And the policy gradient is updated in the following way (without considering the PPO's clipping for simplicity)
\begin{align*}
     &\begin{aligned}
         \sum_{t=0}^{T-1}\nabla_{\theta}\log\pi_{\theta}(a_t|s_t)G_t  &= \sum_{t=0}^{T-1}\nabla_{\theta}  \left( \sum_{d\in D} \log\pi_{\theta}(a^{d}_{t}|s_t) \right)G_t\\
         &= \sum_{t=0}^{T-1}\nabla_{\theta}  \log \left( \prod_{d\in D} \pi_{\theta}(a^{d}_{t}|s_t) \right)G_t
     \end{aligned} \\
     &D = \{\text{Source Unit},\text{Action Type},\text{Move Parameter},\text{Harvest Parameter},\text{Return Parameter},\\
     &\text{Produce Direction Parameter},\text{Produce Type Parameter},\text{Attack Target Unit},\}
\end{align*}

Implementation wise, for each Action Component of range $[0, x-1]$, the logits of the corresponding shape $x$ is generated, which we call Action Component logits, and each $a^{d}_{t}$ is sampled from this Action Component logits. Because of this idea, the algorithm now only has to generate $hw+6+4+4+4+4+6+hw = 2hw + 36$ number of logits, which is significantly less than $9216(hw)^2$. To the best of our knowledge, this approach of handling large multi discrete action space is only mentioned by Kanervisto et, al~\cite{kanervisto2020action}.

\subsubsection{Invalid Action Masking}
Invalid action masking is a technique that ``masks out'' invalid actions and then just sample from those actions that are valid~\citep{vinyals2017starcraft,Berner2019Dota2W}. \cite{huang2020closer} show invalid action masking is crucial in helping the agents explore in $\mu$RTS

\begin{table}[t]
\centering
\caption{The list of hyperparameters and their values.}
\begin{tabular}{ll} 
\toprule
Parameter Names  & Parameter Values\\
\midrule
Total Time Steps & 1,000,000  \\ 
Number of Mini-batches & 4 \\
Number of Environments & 8 \\
Number of Steps per Environment & 128 \\
$\gamma$ (Discount Factor) & 0.99 \\ 
$\lambda$ (for GAE) & 0.95 \\ 
$\varepsilon$ (PPO's Clipping Coefficient) & 0.1 \\ 
$\eta$ (Entropy Regularization Coefficient) & 0.01 \\ 
$\omega$ (Gradient Norm Threshold)& 0.5 \\
$K$ (Number of PPO Update Iteration Per Epoch)& 4 \\
$\alpha$ Learning Rate &  0.00025 Linearly Decreased to 0 \\
& over the Total Time Steps\\
$c1$ (Value Function Coefficient, see Equation~\ref{eq:full-objective})& 0.5\\
$c2$ (Entropy Coefficient, see Equation~\ref{eq:full-objective})& 0.01\\
\bottomrule
\end{tabular}
\label{tab:params}
\end{table}  

\subsubsection{Code-level Optimizations}
Here is a list of code-level optimizations utilized in this experiments. For each of these optimizations, we include a footnote directing the readers to the files in the  \emph{openai/baselines}~\citep{baselines} that implements these optimization.

\begin{enumerate}
    \item \textbf{Normalization of Advantages\footnote{\url{https://github.com/openai/baselines/blob/ea25b9e8b234e6ee1bca43083f8f3cf974143998/baselines/ppo2/model.py\#L139}}:} After calculating the advantages based on GAE, the advantages vector is normalized by subtracting its mean and divided by its standard deviation.
    \item \textbf{Normalization of Observation\footnote{\url{https://github.com/openai/baselines/blob/ea25b9e8b234e6ee1bca43083f8f3cf974143998/baselines/common/vec_env/vec_normalize.py\#L4}}:} The observation is pre-processed before feeding to the PPO agent. The raw observation was normalized by subtracting its running mean and divided by its variance; then the raw observation is clipped to a range, usually $[-10,10]$.
    \item \textbf{Rewards Scaling\footnote{\url{https://github.com/openai/baselines/blob/ea25b9e8b234e6ee1bca43083f8f3cf974143998/baselines/common/vec_env/vec_normalize.py\#L4}}:} Similarly, the reward is pre-processed by dividing the running variance of the discounted the returns, following by clipping it to a range, usually $[-10,10]$.
    \item \textbf{Value Function Loss Clipping\footnote{\url{https://github.com/openai/baselines/blob/ea25b9e8b234e6ee1bca43083f8f3cf974143998/baselines/ppo2/model.py\#L68-L75}}:} The PPO implementation of {\em  openai/baselines} clips the value function loss in a manner that is similar to the PPO's clipped surrogate objective:

    \[V_{loss} =\max \left[\left(V_{\theta_{t}}-V_{t a r g}\right)^{2},\left(V_{\theta_{t-1}} + \mbox{clip}\left(V_{\theta_{t}}-V_{\theta_{t-1}}, -\varepsilon, \varepsilon\right)\right)^{2}\right]\]
    where $V_{t a r g}$ is calculated by adding $V_{\theta_{t-1}}$ and the  $A$ calculated by General Advantage Estimation\cite{schulman2015high}.
    \item \textbf{Adam Learning Rate Annealing\footnote{\url{https://github.com/openai/baselines/blob/ea25b9e8b234e6ee1bca43083f8f3cf974143998/baselines/ppo2/ppo2.py\#L135}}:} The Adam \cite{kingma2014adam} optimizer's learning rate is set to decay as the number of timesteps agent trained increase.
    \item \textbf{Mini-batch updates\footnote{\url{https://github.com/openai/baselines/blob/ea25b9e8b234e6ee1bca43083f8f3cf974143998/baselines/ppo2/ppo2.py\#L160-L162}}:} The PPO implementation of the {\em  openai/baselines} also uses minibatches to compute the gradient and update the policy instead of the whole batch data such as in {\em open/spinningup}.
    The mini-batch sampling scheme, however, still makes sure that every transition is sampled only once, and that the all the transitions sampled are actually for the network update. 
    \item \textbf{Global Gradient Clipping\footnote{\url{https://github.com/openai/baselines/blob/ea25b9e8b234e6ee1bca43083f8f3cf974143998/baselines/ppo2/model.py\#L107}}:} For each update iteration in an epoch, the gradients of the policy and value network are clipped so that the ``global $\ell_{2}$ norm'' (i.e. the norm of the concatenated gradients of all parameters) does not exceed 0.5.
    \item \textbf{Orthogonal Initialization  of weights\footnote{\url{https://github.com/openai/baselines/blob/ea25b9e8b234e6ee1bca43083f8f3cf974143998/baselines/a2c/utils.py\#L58}}:} The weights and biases of fully connected layers use with orthogonal initialization scheme with different scaling. For our experiments, however, we always use the scaling of 1 for historical reasons. 
\end{enumerate}

\subsubsection{Neural Network Architecture}
 The input to the neural network is a tensor of shape $(10, 10, 27)$. The first hidden layer convolves 16 $3\times3$ filters with stride 2 with the input tensor followed by  a  rectifier nonlinearity\cite{nair2010rectified}. The second hidden layer similarly convolves 32 $2\times2$ filters  followed by a  rectifier nonlinearity. The final hidden layer is a fully connected linear layer consisting of 128 rectifier units. The policy's output layer is a fully connected linear layer with $2hw + 36=236$ number of output and the value output layer is a fully connected linear layer with a single output.

\subsection{Learning Curves}
\label{sec:learningcurves}
\begin{figure}[t]

  \centering
   \includegraphics[width=\textwidth]{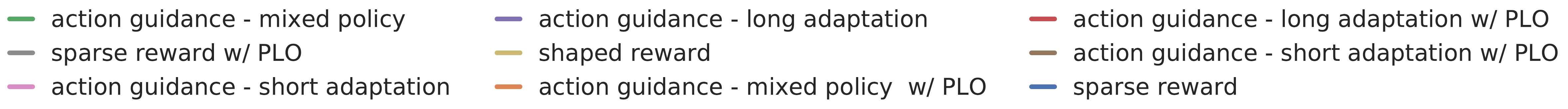}
     \begin{subfigure}[b]{0.32\textwidth}
         \centering
         \includegraphics[width=\textwidth]{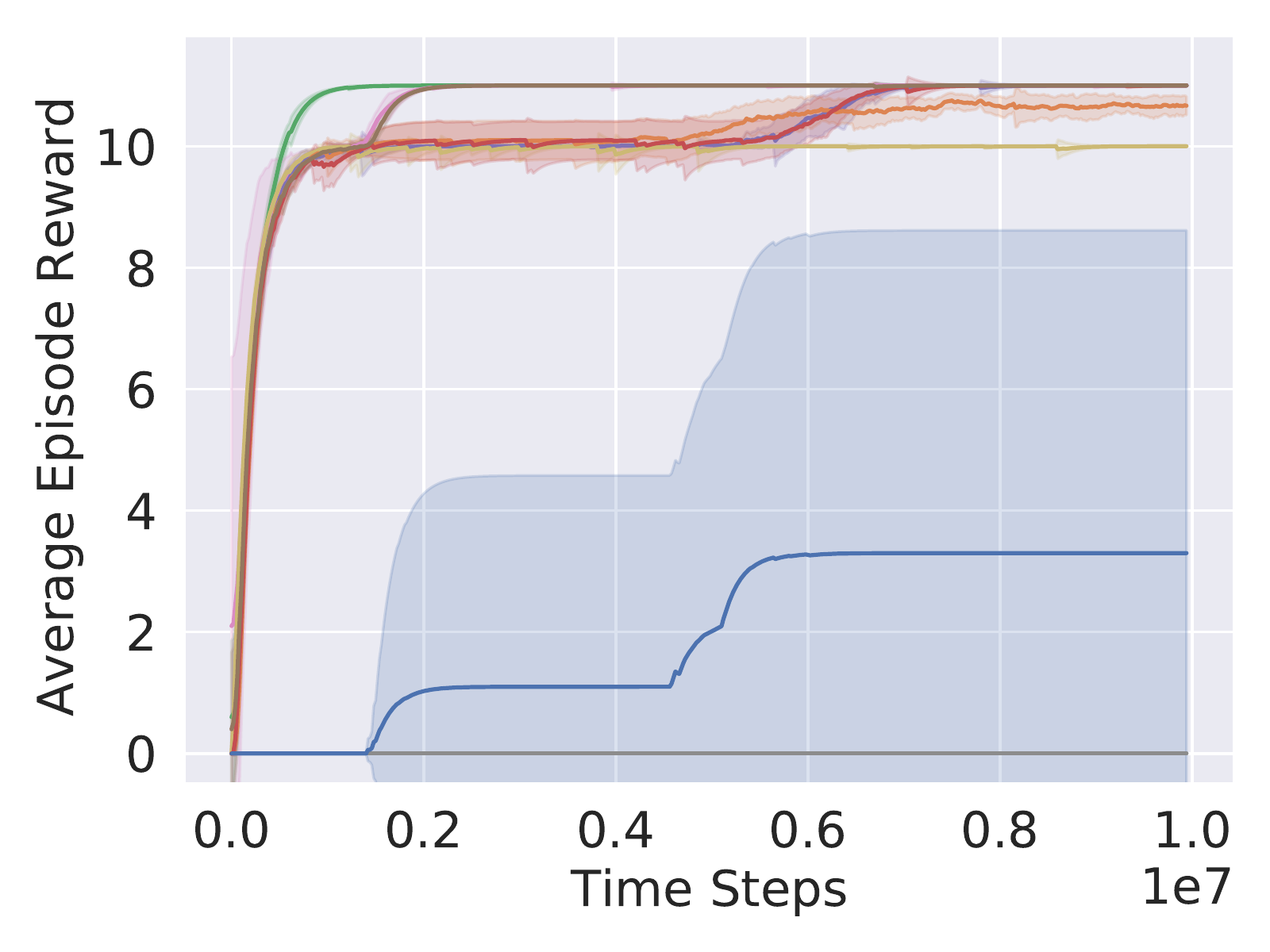}
         \caption{LearnToAttack}
     \end{subfigure}
     \begin{subfigure}[b]{0.32\textwidth}
         \centering
         \includegraphics[width=\textwidth]{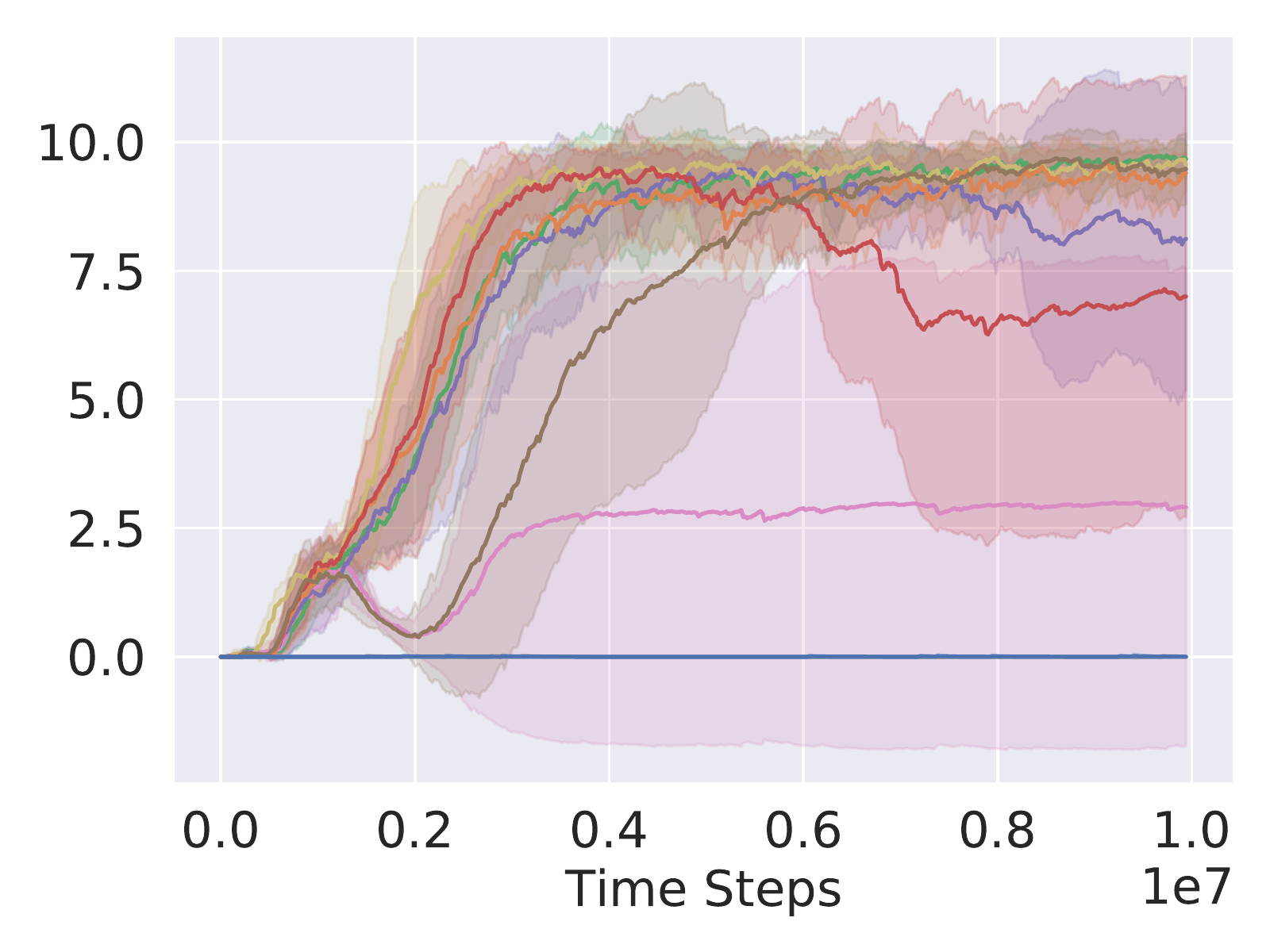}
         \caption{ProduceCombatUnit}
     \end{subfigure}
     \begin{subfigure}[b]{0.32\textwidth}
         \centering
         \includegraphics[width=\textwidth]{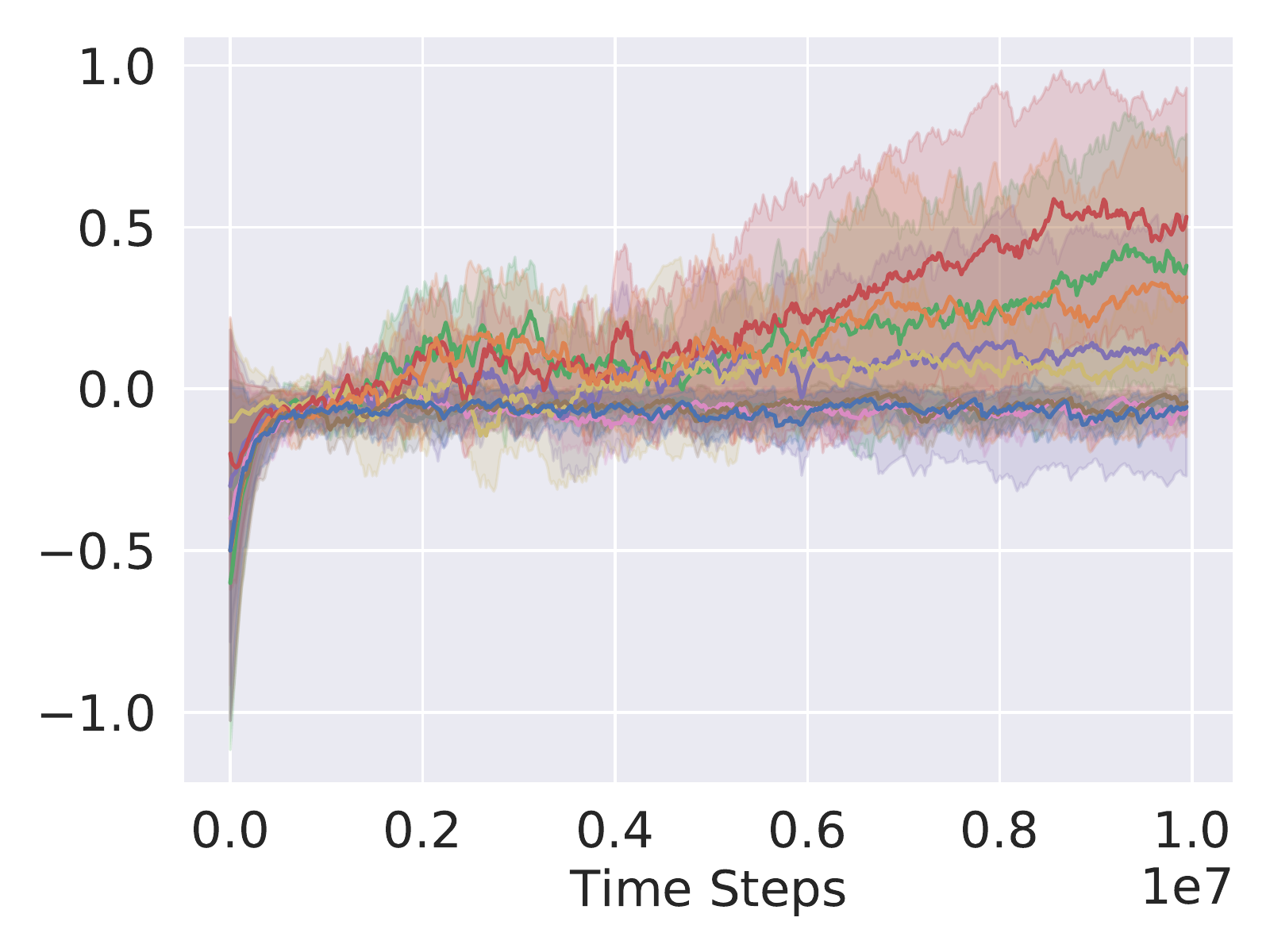}
         \caption{DefeatRandomEnemy}
     \end{subfigure}
       \caption{The learning curves of agents. The x-axis shows the number of time steps and y-axis shows the average episode reward gathered.}
      \label{fig:allLearningCurves}
\end{figure}

The learning curves of all experiments are shown in Figure~\ref{fig:allLearningCurves}.

\end{document}

%% file: math_commands.tex
\usepackage{amsmath,amsfonts,bm}

\def\eqref#1{equation~\ref{#1}}

\def\1{\bm{1}}

\DeclareMathAlphabet{\mathsfit}{\encodingdefault}{\sfdefault}{m}{sl}
\SetMathAlphabet{\mathsfit}{bold}{\encodingdefault}{\sfdefault}{bx}{n}

